\documentclass[11pt,letterpaper]{article}

\usepackage{mathptmx} 
\usepackage[left=1in,right=1in,top=1in,bottom=1in]{geometry}

\usepackage{amsmath}
\usepackage{amsfonts}
\usepackage{amssymb}
\usepackage{graphicx}

\usepackage{placeins}		
\usepackage{enumitem}		
\usepackage{color}			
\usepackage[normalem]{ulem}

\usepackage[numbers,sort&compress]{natbib} 
\makeatother
\usepackage{hyperref}
\usepackage{booktabs}
\usepackage{makecell}
\usepackage[hang,flushmargin]{footmisc}

\newcommand{\bl}[1]{\textcolor[rgb]{0.60,0.75,1}{#1}}

\newcommand*{\TitleFont}{%
      \usefont{\encodingdefault}{\rmdefault}{b}{n}%
      \fontsize{18}{20}%
      \selectfont}
      
\font\Authfont=cmr12 at 14pt
\font\Affilfont=cmr12 at 11pt

\def\correspondingauthor{\footnote{Corresponding author.\newline 
 \hspace*{1.8em}Email: $zzhan59@lsu.edu$}}

\title{\TitleFont Condition Assessment of Stay Cables through Enhanced Time Series Classification Using a Deep Learning Approach}
\author{\Authfont Zhiming Zhang$^{1}$\correspondingauthor{}, Jin Yan$^{2}$, Liangding Li$^{3}$, Hong Pan$^{4}$, Chuanzhi Dong$^{5}$,} 

\date{}

\usepackage{graphicx}
\usepackage{subcaption}
\graphicspath{ {Figures/} }
\begin{document}

\begin{minipage}{1\textwidth}\raggedleft
\bl{The 1st International Project Competition for Structural Health Monitoring}\\
\bl{IPC-SHM, 2020}\\
\bl{June 15 - September 30, 2020, Harbin, China}
\end{minipage}

{\let\newpage\relax\maketitle}

\vspace{-1cm}

\begin{center}
\Affilfont{\textit{$^{1}$ Department of Civil and Environmental Engineering, Louisiana State University, Baton Rouge, USA\newline
$^{2}$Department of Civil, Construction and Environmental Engineering, Iowa State University, Ames, USA \newline
$^{3}$ Department of Computer Science, University of Central Florida, Orlando, USA \newline
$^{4}$ Department of Civil and Environmental Engineering, North Dakota State University, Fargo, USA \newline
$^{5}$  Department of Civil, Environmental, and Construction Engineering, University of Central Florida, Orlando, USA
\newline}}
\end{center}

	\section*{ABSTRACT}
	Stay cables  play an essential role on cable-stayed bridges. Severe vibrations and/or harsh environment may result in cable failures. Hence, an efficient structural health monitoring (SHM) solution for cable damage detection is necessary. To this end, the present study proposes a data-driven method that detects cable damage from measured cable forces by recognizing biased patterns from the intact conditions. The proposed method solves the pattern recognition problem for cable damage detection through time series classification (TSC) in deep learning, considering that the cable's behavior can be implicitly represented by the measured cable force series. A deep learning model, long short term memory fully convolutional network (LSTM-FCN), is leveraged by assigning appropriate inputs and representative class labels for the TSC problem,  First, a TSC classifier is trained and validated using the data collected under intact conditions of stay cables, setting the segmented data series as input and the cable (or cable pair) ID as class labels. Subsequently, the classifier is tested using the data collected under possible damaged conditions. Finally, the cable or cable pair corresponding to the least classification accuracy is recommended as the most probable damaged cable or cable pair. The proposed method was tested on an in-service cable-stayed bridge with damaged stay cables. Two scenarios in the proposed TSC scheme were investigated: 1) raw time series of cable forces were fed into the classifiers; and 2) cable force ratios were inputted in the classifiers considering the possible variation of force distribution between cable pairs due to cable damage. Combining the results of TSC testing in these two scenarios, the cable with rupture was correctly identified. This study proposes a data-driven methodology for cable damage detection that requires the least data preprocessing and feature engineering, which enables fast and convenient early detection in real applications. 

	\section*{INTRODUCTION}\label{Sec:Intro}
	
	Stay cables are among the most critical elements of cable-stayed bridges, since they provide essential support to the bridge deck \cite{li2018condition}. Stay cables are susceptible to fatigue and corrosion damage, and cable deterioration or failure on large span bridges have been reported in the literature \cite{li2016state,brownjohn2011vibration,ko2005technology}.
	When wire rupture occurs to a certain cable, it leads to redistribution of dead and live loads that are applied to other cables and thus can endanger the whole bridge (\cite{macdonald2005variation,miyashita2008vibration}. Therefore, the condition assessment of stay cables is of critical importance for cable-stayed bridges \cite{li2014design,yang2016real}. Cable tension monitoring is one of the widely used approaches to evaluate the health condition of bridge cables \cite{li2014real}, which can be fulfilled by either monitoring the strain of cable wires with strain sensors or assessing the total cable forces with load cell, accelerometers, etc. \cite{li2014real,yang2016real,li2009applications}
	
	One of the challenges in stay cable damage detection is that the results of cable tension analysis do not directly indicate the health condition of stay cables. The measured cable tension in field tests is not only affected by the cable condition but also the traffic loading magnitude, environmental factors (wind, temperature, moisture, etc.), and transmission noises of sensors \cite{li2016state}. For example, many traditional methods evaluate the loss of cable tension and damage condition using the variation of natural frequencies \cite{miyashita2008vibration}, which is largely affected by the external factors mentioned above. Therefore, it necessitates developing a health indicator that is sensitive to the cable's health condition and insusceptible to those exterior factors. Li et al. used the Gaussian mixture model (GMM) to simulate the measured cable tension ratios between cable pairs and evaluated the cable conditions from the observed transition of model patterns \cite{li2016state}. However, this approach requires intensive data preprocessing, for example clustering corresponding to traffic lanes and source separation, and the used machine learning model is considerably more complex than a widely used deep learning model in the literature.
	
	Pattern recognition and machine learning have been widely used in SHM since the beginning of this decade. Applications include data compression and recovery, anomaly detection, knowledge discovery for structural condition assessment, etc.\cite{bao2019state,sun2020review}. For example, deep learning models such as convolutional neural network (CNN) and recurrent neural network (RNN) have been widely used for time series analysis in SHM. CNN has been used to analyze measured acceleration series for bolt loosening detection \cite{abdeljaber20181} and structural damage localization \cite{wang2009review,lin2017structural}. RNN including the long short-term memory (LSTM) or gated recurrent unit (GRU) has been used for remaining life estimation \cite{zheng2017long} and degradation assessment \cite{guo2017recurrent} of machines and response prediction in structural dynamics\cite{zhang2020physics}. It is noted that in SHM, most collected data (e.g., acceleration, strain, temperature, etc.) are in the form of time series, which need to be further processed and analyzed to evaluate possible structural damage or deficiency. 
	
	Time series classification (TSC) has been widely used for data mining and knowledge discovery in various areas, such as health care \cite{abdelfattah2018augmenting,ma2018health,fawaz2018evaluating}, financial analysis \cite{higdon2019time,chao2019novel}, and weather forecast \cite{gao2019day}. TSC methods include feature-based approaches based on non-deep machine learning models \cite{orsenigo2010combining,seto2015multivariate,xingbrief} and deep learning approaches \cite{fawaz2019deep,zheng2014time}. Compared with feature-based approaches, deep learning approaches obviate the demanding work of feature extraction and feature engineering. Comprehensive reviews of deep learning for TSC can be found in \cite{fawaz2019deep,santos2016literature,langkvist2014review}. Karim et al. \cite{karim2017lstm,karim2019insights,karim2019multivariate} proposed augmenting the fully convolutional networks (FCN) with the long short term memory recurrent neural network (LSTM-RNN) module for TSC. The proposed model avoids complex data preprocessing and outperforms traditional TSC models in terms of classification accuracy. Experimental studies approve that concatenating the LSTM features with the CNN features improves the robustness of learned TSC classifier.
	
	In this study, the condition assessment of stay cables is solved in the framework of TSC using the LSTM-FCN model. Assigning the collected time series data with class labels indicating their sources, a TSC classifier learned using the intact data will not generalize well to the data collected under damaged condition. A case study with field test data measured from a real cable-stayed bridge validates the proposed methodology. The proposed deep learning approach successfully solves the cable damage detection problem with sufficient accuracy using the designed TSC framework. It obviates the intensive work for data preprocessing or feature engineering without sacrifice of identification accuracy, which improves its convenience and  efficiency in practical applications on real bridges. The rest of this paper will first elaborate the methodology proposed in this study and then present an experimental case study on an in-service cable-stayed bridge. Finally, concluding remarks will be presented.
	
	\section*{METHODOLOGY}
\begin{figure*}[!h]
		\centering
		\includegraphics[scale=0.55]{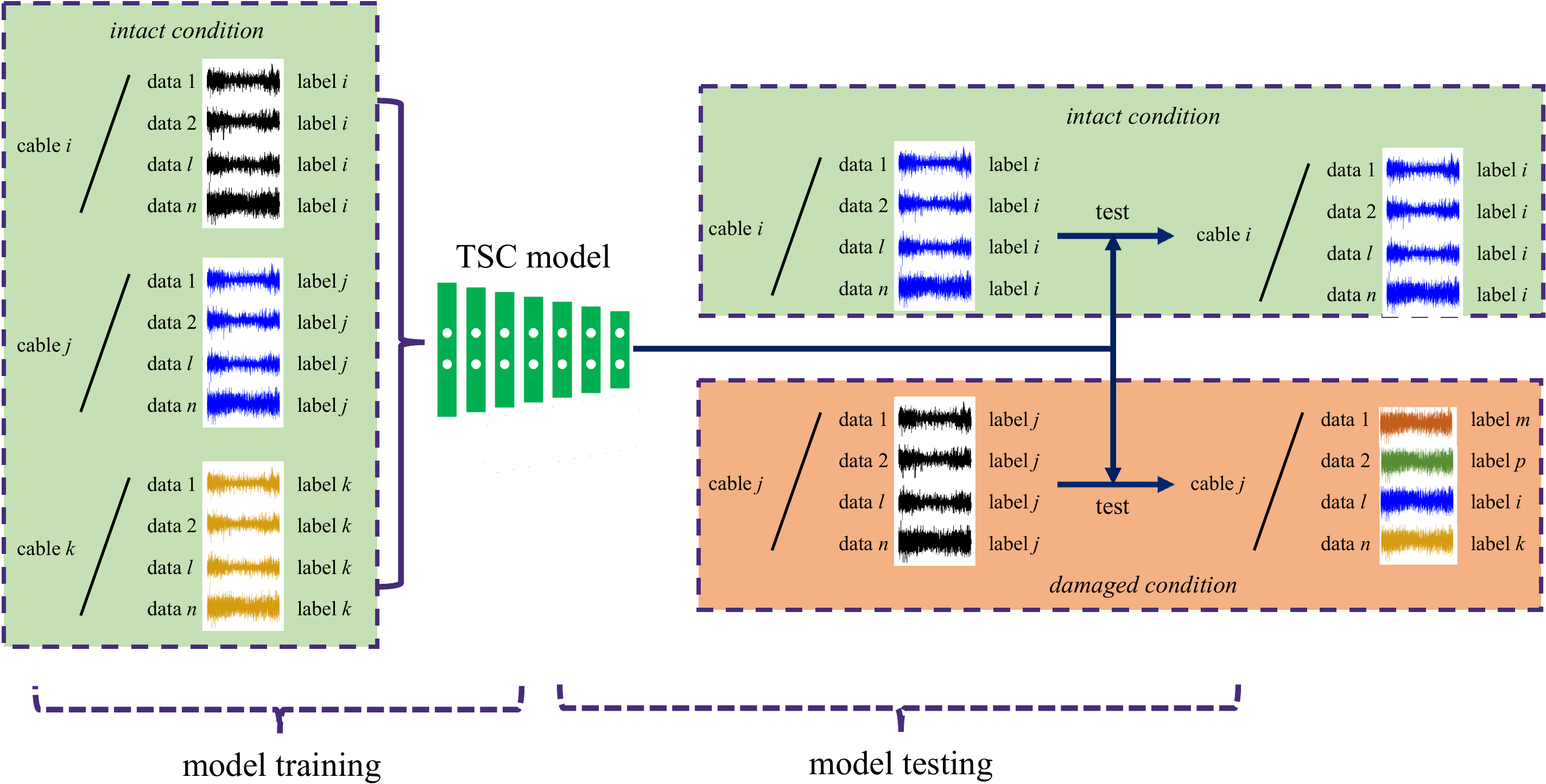}
		\caption{The idea of TSC for condition assessment of stay cables.}
		\label{Figure:idea}
	\end{figure*}

	In this study, the conditions of stay cables are evaluated through TSC with deep learning approach, with the idea as shown in Figure \ref{Figure:idea}. Under intact condition, a large amount of time series data can be collected from each instrumented cable. The measured quantities can be accelerations, strains, temperatures, etc. Since all the collected data are documented with the cable ID from which they are measured, a multi-class TSC problem can be formulated by setting the cable ID of a certain time series of measured quantity as the target label. Hence, a classifier can be learned using the data collected under intact condition, which should be sufficiently robust to account for the variations of environmental factors and traffic situation (see Figure \ref{Figure:idea} (a)). When rupture happens to a certain cable, transition occurs to the state of measured time series from that cable, which makes their classes hard to correctly identify using the learned classifier (see Figure \ref{Figure:idea} (b)). Therefore, the cable(s) with non-trivial rupture can be detected by identifying the one(s) with measured data that yield abnormally low classification accuracy when tested using the learned classifier. 
	
	For a certain cable, Let $X = [X^1, X^2 ,..., X^M]$ denote an $M$-dimensional  measured time series sample, with the superscript $M$ being the number of measured quantities and $X^i \in \mathbb{R}$, $i = 1,2,...M$. Then a dataset $D = \left\lbrace(X_1,Y_1),(X_2,Y_2),...,(X_N,Y_N)\right\rbrace$ denotes a set of sample pairs $\left(X_i,Y_i\right)$ for all measured data segments with $X_i$ denoting a univariate ($M=1$) or multivariate ($M>1$) time series, $Y_i$ its corresponding cable ID, and $N$ the number of segmented data series. With the dataset prepared, the task of TSC is learning an efficient classifier that maps the input $X$ to its probability distribution over all possible classes with sufficient accuracy. Subsequently, in the testing stage, the newly collected data $D^t= \left\lbrace(X_1,Y_1),(X_2,Y_2),...,(X_{Nt},Y_{Nt})\right\rbrace$ on a later date are inputted into the previously learned classifier. Since all the measured data are accompanied with the cable IDs they belong to, their class labels are assumed known and are used to evaluate the classification accuracy. If the average output classification accuracy of the time series data collected from a certain cable is significantly lower than other cables and/or the average classification accuracy with intact data, it can be determined that rupture or deficiency has been detected on that cable. It should be noted that preprocessing is necessary to eliminate the data collected from malfunctioning sensors to avoid misidentification. 
	
\begin{figure*}[!h]
		\centering
		\includegraphics[scale=0.4]{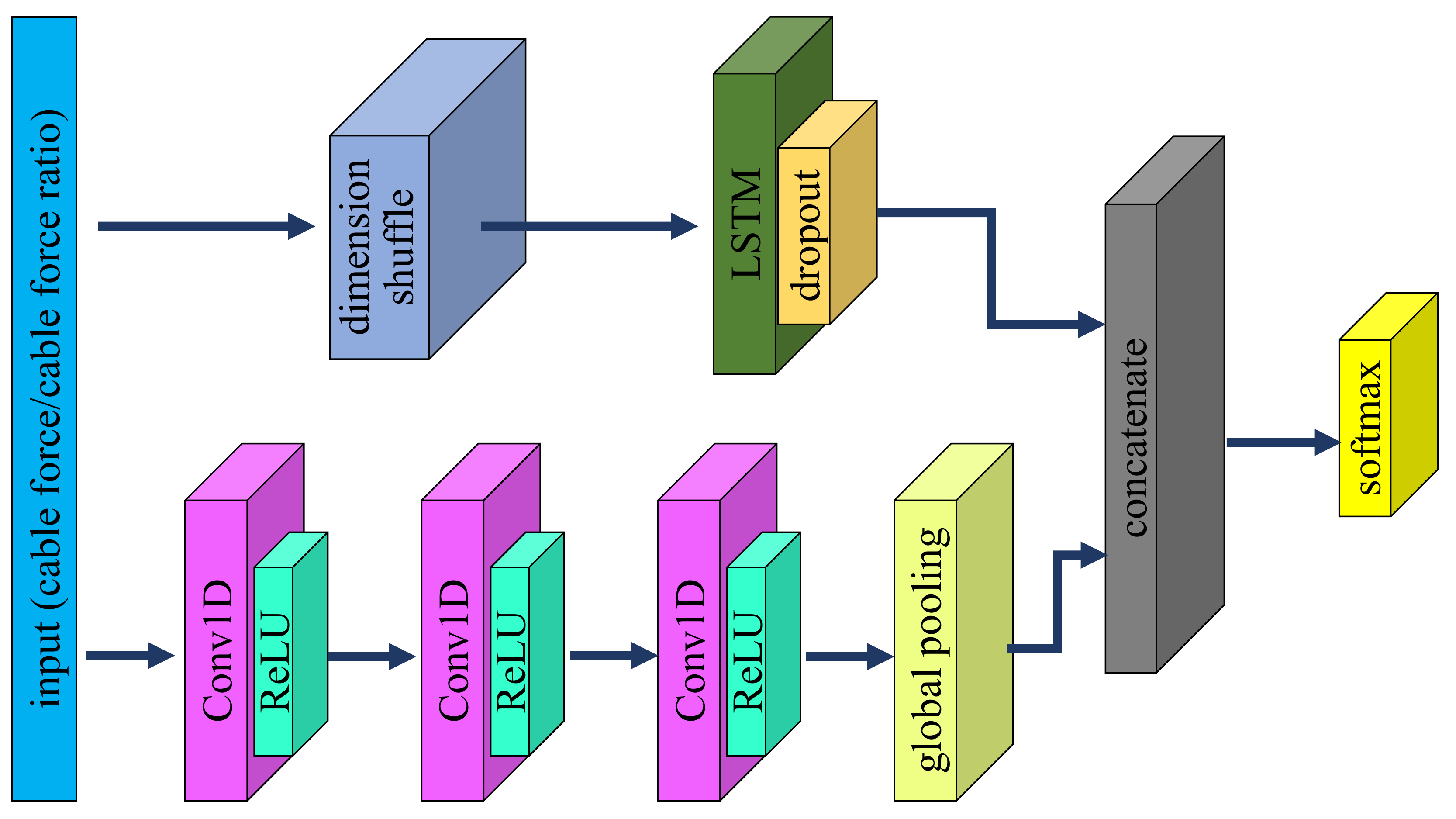}
		\caption{Architecture of LSTM-FCN (reproduced from \protect{\cite{karim2017lstm}}).}
		\label{Figure:LSTM-FCN}
	\end{figure*}

	The LSTM-FCN model is used for TSC in this cable condition assessment problem considering its proven advantages over other TSC models \cite{karim2017lstm,karim2019insights,karim2019multivariate}. Figure \ref{Figure:LSTM-FCN} illustrates the architecture of LSTM-FCN model for univariate TSC that is used in the experimental analysis of this study. The input can contain time series of any parameters that are evaluated as contributive to assessing the cable condition. The cable forces or their ratios are used in this study, and the details with be elaborated in \textbf{EXPERIMENTAL STUDY}. 
	
	The deep learning TSC model used in this study consists of an FCN module and an LSTM module. The FCN module contains three stacked temporal convolutional blocks, and the kernel sizes will be determined in \textbf{EXPERIMENTAL STUDY}. Each temporal convolutional block consists of a temporal convolutional layer for feature extraction and a ReLU activation function. Finally, global average pooling is imposed on the output of the final convolutional block to reduce the number of parameters in the FCN model and thus decreases its complexity.
	
	Additionally, the LSTM module is used to augment the feature vector obtained from FCN, which acts as a regularizer to FCN and thus is expected to improve its performance. In the LSTM module, the input time series is dimension-shuffled, which transforms the input to a multivariate single step vector. This dimension shuffle has the potential of reducing the rapid overfitting issue of the LSTM model and improving the efficiency of model training. The LSTM block following the dimension shuffle is comprised of an LSTM layer with dropout. Finally the concatenated features are passed to a softmax layer for multiclass classification.
	
	\section*{EXPERIMENTAL STUDY}
	\label{S:Exp}
\begin{figure*}[!h]
		\centering
		\includegraphics[scale=1.0]{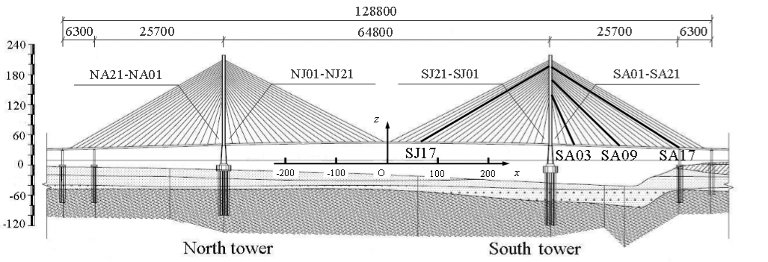}
		\caption{The investigated cable-stayed bridge \protect{\cite{li2018condition}}. The unit is m for all dimensions.}
		\label{Figure:Bridge}
	\end{figure*}

	In this section, an experimental study is implemented to validate the proposed deep learning framework for condition assessment of stay cables. The investigated bridge is a large-span cable-stayed bridge in Mainland China \cite{li2018condition}, with a main span of 648 m and two side spans of 63 + 257 m, as shown in Figure \ref{Figure:Bridge}. All the cables (168 in total) were installed with anchorage load cells, which have a sampling frequency of 2 Hz. Cables are numbered according to their locations as shown in Figure \ref{Figure:Bridge}. ``N" and ``S" denote the north and south side, respectively, that is, the left and right side in the figure; ``J" and ``A" denote the main and side span, respectively; ``U" and ``D" denote the upriver and downriver sides, respectively. For example, SJ11 represents the $11^\mathrm{th}$ cable pair on the south side span, and it contains SJU11 on the upriver side and SJD11 on the downriver side. The bridge has been open to the public since October 2005.
	
	In this study, 14 cables (SJS08 to SJS14 and SJX08 to SJX14) are investigated for stay cable condition assessment, and their tension forces in 10 days (2006-05-13 to 2006-05-19, 2007-12-14, 2009-05-05, and 2011-11-01) are available for analysis and decision making. All of the 14 cables were intact prior to the year 2011, and rupture of wires occurred to a certain cable in 2011. The task of this experimental project is to identify the damaged cable using the collected data.
	
	\FloatBarrier
	\subsection*{Data Preprocessing}
\begin{figure*}[!h]
		\centering
		\includegraphics[scale=0.65]{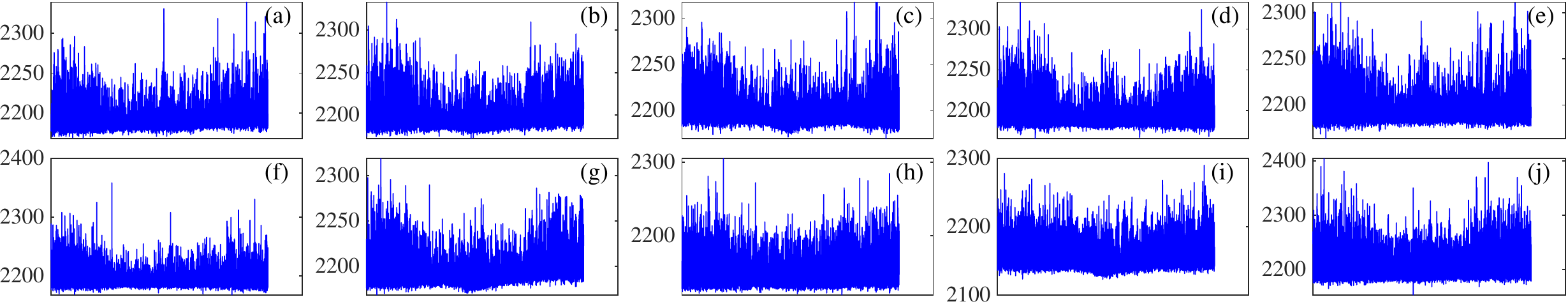}
		\caption{Cable force of SJS08 with the unit kN. (a) to (j) show the data on the following ten days respectively: 2006-05-13 to 2006-05-19, 2007-12-14, 2009-05-05, and 2011-11-01.}
		\label{Figure:cf_SJS08}
	\end{figure*}

\begin{figure*}[!h]
		\centering
		\includegraphics[scale=0.65]{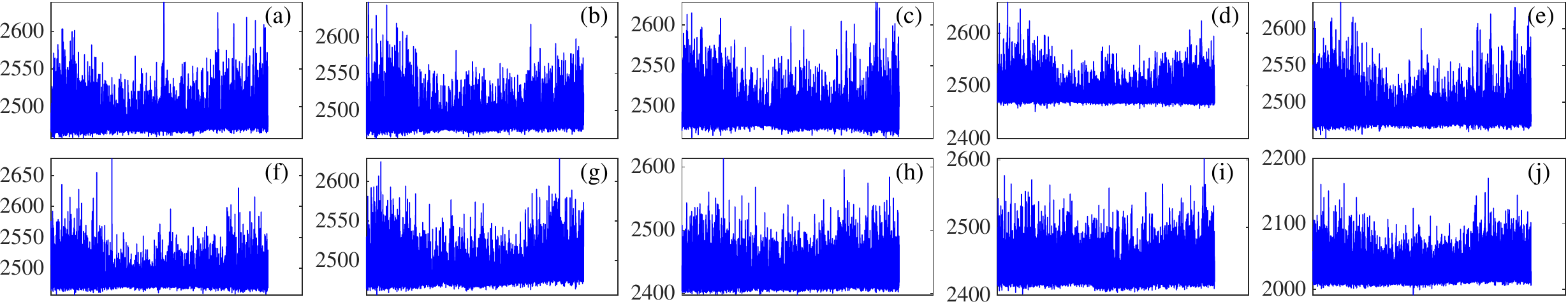}
		\caption{Cable force of SJS11 with the unit kN. (a) to (j) show the data on the following ten days respectively: 2006-05-13 to 2006-05-19, 2007-12-14, 2009-05-05, and 2011-11-01.}
		\label{Figure:cf_SJS11}
	\end{figure*}

\begin{figure*}[!h]
		\centering
		\includegraphics[scale=0.65]{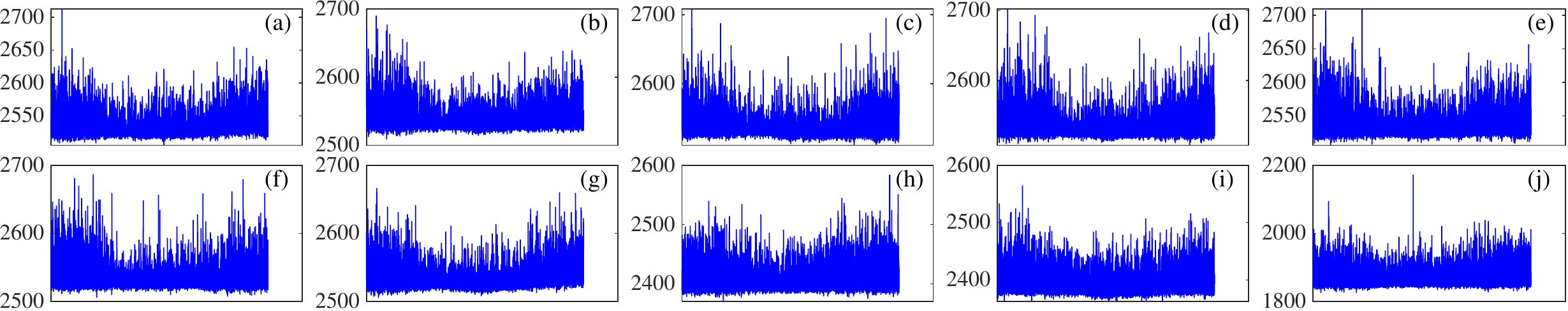}
		\caption{Cable force of SJX11 with the unit kN. (a) to (j) show the data on the following ten days respectively: 2006-05-13 to 2006-05-19, 2007-12-14, 2009-05-05, and 2011-11-01.}
		\label{Figure:cf_SJX11}
	\end{figure*}

	Figures \ref{Figure:cf_SJS08} to \ref{Figure:cf_SJX11} show the measured tension forces from three example cables (i.e., SJS08,SJS11, and SJX11) in the ten days. All cable forces show a strong daily periodicity pattern probably due to changes in traffic loads or environmental conditions such as temperature. In data preprocessing, extreme values exceeding the thresholds of cable forces are excluded from the data record. 
	
\begin{figure*}[!h]
		\centering
		\includegraphics[scale=0.65]{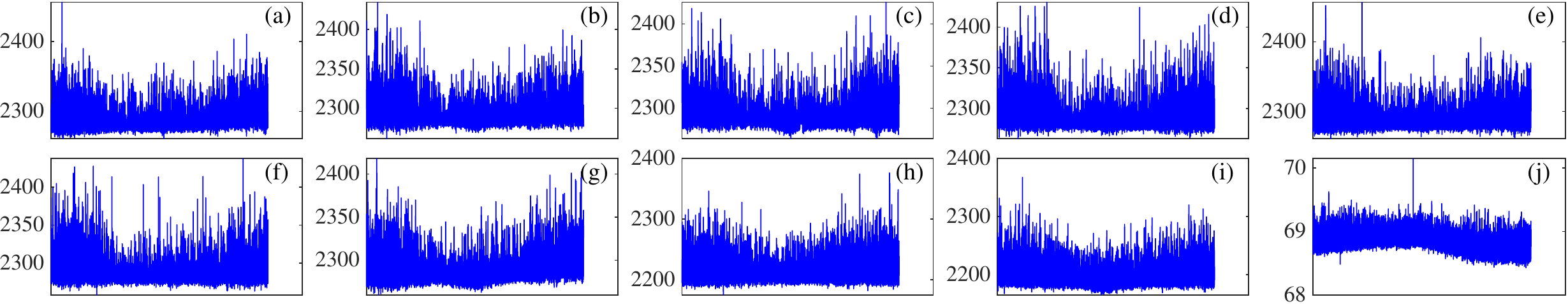}
		\caption{Cable force of SJX08 with the unit kN. (a) to (j) show the data on the following ten days respectively: 2006-05-13 to 2006-05-19, 2007-12-14, 2009-05-05, and 2011-11-01.}
		\label{Figure:cf_SJX08}
	\end{figure*}

\begin{figure*}[!h]
		\centering
		\includegraphics[scale=0.65]{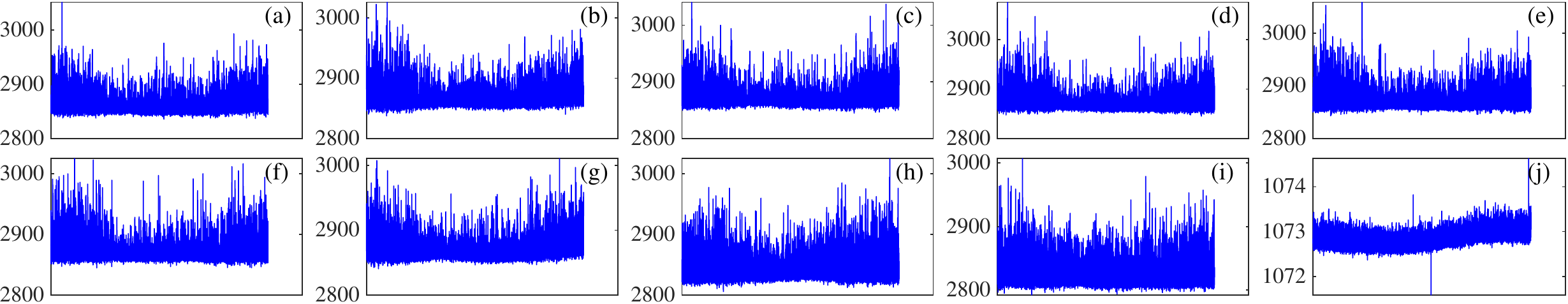}
		\caption{Cable force of SJX13 with the unit kN. (a) to (j) show the data on the following ten days respectively: 2006-05-13 to 2006-05-19, 2007-12-14, 2009-05-05, and 2011-11-01.}
		\label{Figure:cf_SJX13}
	\end{figure*}

	Figure \ref{Figure:cf_SJX08} shows the cable force records on SJX08. It can be observed that on the day 2011-11-01, the recorded cable force is notably small (see figure (j)) compared with that on prior dates. Moreover, the amplitude fluctuation is within 2 kN on that day, which is much smaller than that of previous dates (i.e., approximately 100 kN). This abnormal pattern indicates possible cable damage or sensor failure. Moreover, the cable force exhibits a reversal or random drift pattern that does not change with the traffic loads or environmental conditions. Therefore, it  can be determined that the load cell on SJX08 has failed prior to 2011-11-01. A similar phenomenon is observed on the cable SJX13, as shown in Figure \ref{Figure:cf_SJX13} (j). Therefore, these two cables are excluded from cable damage detection. However, the data measured on these two cables under intact conditions are still used in learning a TSC classifier.
	
	When wire rupture occurs in a certain cable, it will cause redistribution of cable forces among the cables located close to that cable, especially the one within the same cable pair. For example, when ruptures occurs to certian wires in cable SJS08, more forces will be imposed on SJX08 when the same vehicle crosses this cable pair on the same lane, assuming identical properties of other parts of the bridge and environmental conditions. Li et al. \cite{li2018condition} has shown that the cable force ratio within a certain cable pair is an efficient indicator of cable condition. Hence, this study also uses the cable force ratio as the input of the deep learning model and will compare the results with that using the raw data as the input.
	
\begin{figure*}[!h]
		\centering
		\includegraphics[scale=0.65]{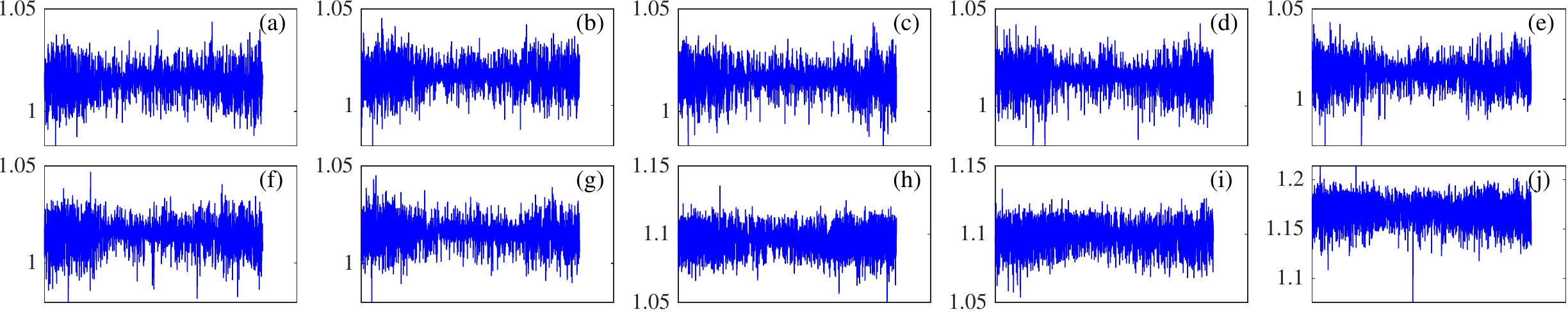}
		\caption{Cable force ratio of SJS09. (a) to (j) show the data on the following ten days respectively: 2006-05-13 to 2006-05-19, 2007-12-14, 2009-05-05, and 2011-11-01.}
		\label{Figure:cfr_SJ9}
	\end{figure*}

\begin{figure*}[!h]
		\centering
		\includegraphics[scale=0.65]{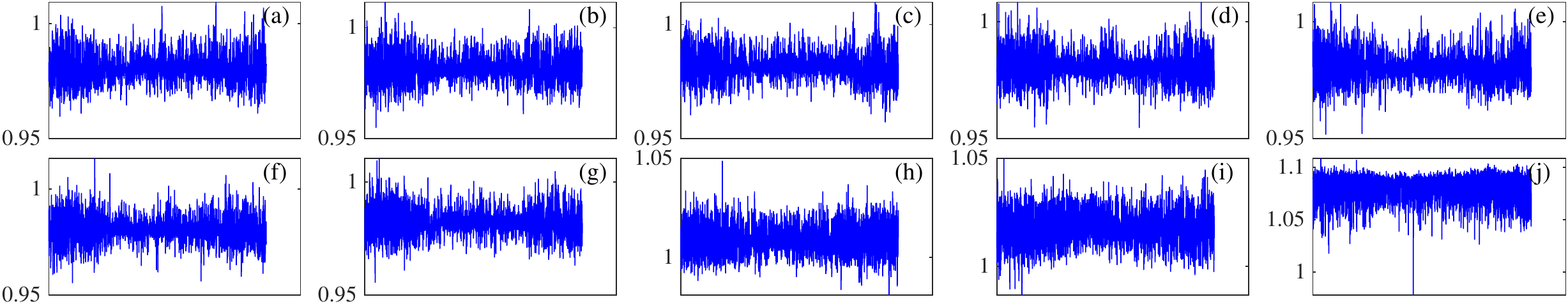}
		\caption{Cable force ratio of SJS11. (a) to (j) show the data on the following ten days respectively: 2006-05-13 to 2006-05-19, 2007-12-14, 2009-05-05, and 2011-11-01.}
		\label{Figure:cfr_SJ11}
	\end{figure*}

	Figures \ref{Figure:cfr_SJ9} and \ref{Figure:cfr_SJ11} show the cable force ratios of two example cable pairs (SJ9 and SJ11). The cable force ratio of certain cable pair is defined as the cable force of the upriver cable divided by that of the downriver cable. It can be observed that all the cable force ratios fluctuate around a certain value close to 1.0, since the force distribution between the two cables in a pair is approximately equal when a certain vehicle crosses this cable pair regardless of which lane it travels in. Moreover, Figure \ref{Figure:cfr_SJ11} shows that the times series of cable force ratio appears fluctuating in a different mode in 2011 (see figure (j)) than before, which is not observed in the time series of cable forces (see Figures \ref{Figure:cf_SJS11} and \ref{Figure:cf_SJX11}). It needs to be verified via pattern recognition whether this abnormal cable force ratio indicates a shift of cable condition.
	
	The time series of a certain cable/cable pair on a certain day has a length of 172800. Each of them is divided to the length of 1600 after trial analysis and following the practice of the UCR datasets \cite{dau2019ucr} that are widely used for TSC. Then each time series segment is labeled with the cable or cable pair ID it belongs to. Hence each class contains $108 \times 9 = 972$ times series samples in the training dataset (before 2011) and 108 samples in the testing dataset (in 2011). It is noted that the TSC problem has 14 class labels when taking the cable forces as input (Scenario 1) and 7 class labels when taking the cable force ratios as input (Scenario 2). Hence, the testing results in Scenario 2 cannot reflect the possibly damaged cable but the cable pair it belongs to, and further analysis is necessary to identify the damaged cable within the detected pair. 
	
	\subsection*{LSTM-FCN Model Configuration and Training}
	As shown in Figure \ref{Figure:LSTM-FCN}, The FCN module contains three 1D convolutional layers in the temporal convolutional blocks and the kernel sizes are set as 128, 256, and 128, respectively, after trial analysis in this case study. The optimal number of LSTM cells was determined as 8 by hyperparameters search over a range of 4 cells to 128 cells. The dropout rate is set as high as 80\% to reduce overfitting. 
	
	Regarding the dataset division, in both the two scenarios, the data collected prior to 2011 are first randomly divided into training and testing data of equal size, in order to learn an efficient model for the intact conditions before testing on the data collected in 2011 for damage detection. This setup improves the robustness of decision making in this study compared with directly testing the trained model on the data of 2011.
	
	The number of training epochs is set as 2000 epochs, and the initial batch size is set as 128. All models in this study are trained using the Keras library with the TensorFlow backend. The Adam optimizer is used in the model training with an initial learning rate of 0.001 and a final learning rate of 0.0001, which is reduced by $1/\sqrt[3]{2}$ every 100 epochs if no improvement of the validation performance is observed. The strategy proposed in \cite{he2015delving} is used to initialize all convolutional kernels. The performance of trained models are evaluated using accuracy.
	
	\FloatBarrier
	\subsection*{Results and Discussions}
	
	This section presents and discusses the results of TSC for stay cable condition assessment using the LSTM-FCN deep learning model. Two scenarios in the proposed TSC scheme were investigated: 1) raw time series of cable forces were fed into the classifiers; and 2) cable force ratios were inputted in the classifiers considering the possible variation of force distribution between cable pairs due to cable damage. First, the TSC results of each scenario are presented for each dataset, including the classificaton accuracy on the testing dataset prior to 2011, the classification accuracy on all data in 2011, and the classification accuracy on all time series segments on each cable (pair) in 2011. Finally, the most probable cable with rupture will be determined by combining the recommendations collected from each scenario, considering that the results in these two scenarios are not independent but complementary for the decision making in this study.
	
	\subsubsection*{\textit{Scenario 1 with Cable Force as Input}}
	
	Table \ref{Table:accS1} lists the classification accuracies of different datasets using the learned model with data collected prior to 2011. The testing results of the dataset from each cable are listed in columns 4 to end of the table. Since the cable force magnitude itself is not directly related to the cable properties or damage condition, the classification accuracy is not as high as expected (i.e., 0.79) on data collected under intact conditions prior to 2011. When testing on the unseen data in 2011, the overall accuracy is as low as 0.56, which means that the learned FCN-LSTM model using the intact data cannot generalize well on the data collected on a new date and probably containing damaged conditions. In detail, the classification accuracies on SJS11 and SJX10 are both around 0.25 and considerably lower than that on other cables which are all above 0.50. The classification accuracy on the two cables with malfunctioning sensors (i.e., SJX08 and SJX13) is not applicable. 
	
	\begin{table*}[!h]
		\centering
		\caption{Testing accuracies of different datasets in Scenario 1.}
		\begin{tabular}{p{1.8cm}p{1.8cm}p{2.2cm}p{1.2cm}p{1.2cm}p{1.2cm}p{1.2cm}p{1.2cm}p{1.2cm}p{1.2cm}} \Xhline{2\arrayrulewidth}
			dataset & pre-2011 & overall-2011 & SJS08 & SJS09 & SJS10 & SJS11 & SJS12 & SJS13 & SJS14 \\ \hline
			accuracy & 0.79 & 0.56 & 0.53 & 0.69 & 0.59 & 0.28 & 0.65 & 0.89 & 0.55\\  
			dataset & - & - & SJX08 & SJX09 & SJX10 & SJX11 & SJX12 & SJX13 & SJX14 \\ 
			accuracy & - & - & - & 0.83 & 0.25 & 0.60 & 0.61 & - & 0.60\\ \hline 
			\multicolumn{10}{l}{pre-2011: data prior to the year of 2011; overall-2011: all data in 2011.}\\
		\end{tabular}\label{Table:accS1}
	\end{table*}
	
	From this comparison of classification accuracy between different cables, one can hypothesize that rupture may have happened to one or both of SJS11 and SJX10. It is hard to determine which one is indeed damaged, since damage on a certain cable can significantly affect the behavior of adjacent cables. Especially considering the insufficient overall classification accuracy, further investigation is necessary into the results in Scenario 2.
	
	\subsubsection*{\textit{Scenario 2 with Cable Force Ratio as Input}}
	
	The testing results on cable pairs in Scenario 2 are tabulated in Table \ref{Table:accS2}. The classification accurcy on data collected prior to 2011 is 0.96 and much higher than that in Scenario 1 (i.e., 0.79), which approves that the cable force ratio is an enhanced representation of the cable properties and conditions compared with the raw measured cable force. Moreover, the increased classification accuracy on the overall data in 2011 (from 0.56 in Scenario 1 to 0.75) indicates that the learned model from intact data has improved generality on data collected on a later date.
	
	\begin{table*}[!h]
		\centering
		\caption{Testing accuracies of different datasets in Scenario 2.}
		\begin{tabular}{p{1.8cm}p{1.8cm}p{2.2cm}p{1.2cm}p{1.2cm}p{1.2cm}p{1.2cm}p{1.2cm}p{1.2cm}p{1.2cm}} \Xhline{2\arrayrulewidth}
			dataset & pre-2011 & overall-2011 & SJ08 & SJ09 & SJ10 & SJ11 & SJ12 & SJ13 & SJ14 \\ \hline
			accuracy & 0.96 & 0.75 & - & 0.94 & 0.95 & 0.02 & 0.94 & - & 0.91\\ \hline   
			\multicolumn{10}{l}{pre-2011: data before the year of 2011; overall-2011: all data in 2011.}\\
		\end{tabular}\label{Table:accS2}
	\end{table*}
	
	Regarding the classification performance on each cable pair, the accuracies are all above 0.90 except that on SJ11 (i.e., 0.02), excluding the cable pairs with malfunctioning sensors. This considerable difference in classification accuracy leads to a recommendation with confidence that damage should have happened to the cable pair SJ11. The final decision will be made via combining this finding with that from the analysis in Scenario 1.
	
	\subsubsection*{\textit{Decision Making on the Damaged Cable}}
\begin{figure*}[!h]
		\centering
		\includegraphics[scale=1.0]{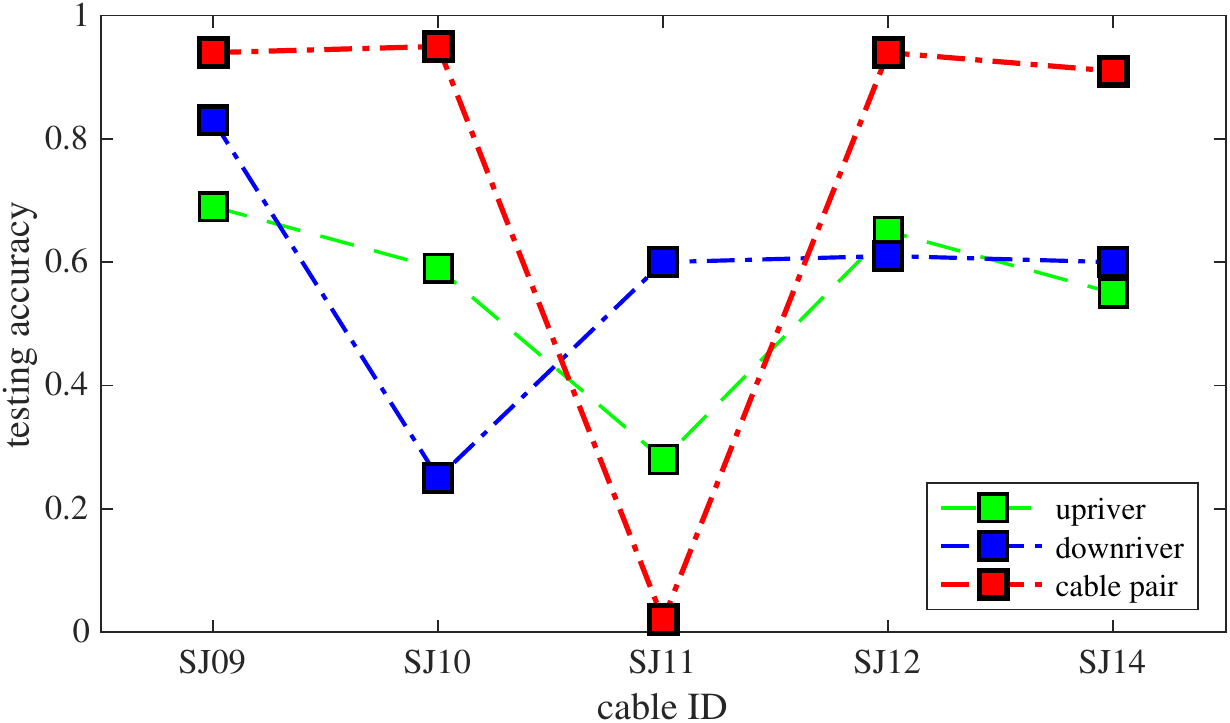}
		\caption{Comparison of classification accuracies in the two scenarios. The curve with legends ``upriver" and ``downriver" are from the results of Scenario 1, and the one with ``cable pair" is from that of Scenario 2.}
		\label{Figure:testAcc}
	\end{figure*}

	Figure \ref{Figure:testAcc} compares the classification accuracy obtained in the two scenarios. Results in Scenario 1 indicate possible damage in SJS11 and SJX10, and results in Scenario 2 indicate damage on the cable pair SJ11. Combining the results in these two scenarios, the most probable damaged cable can be determined as SJS11. This is consistent with the results in \cite{li2018condition}, which detected damage on the same cable pair but failed to identify the exact cable with rupture.This conclusion still needs to be further verified through onsite inspection.
			
	\section*{CONCLUSIONS}
	\label{S:Conclusion}
	Stay cables are the most important component on a cable-stayed bridge which usually has a large span. They are susceptible to fatigue and corrosion damage that may lead to failure of cables and even the whole bridge. Therefore, early detection of cable damage has critical importance for bridge maintenance. To this end, this study proposes detecting cable damage from collected cable force data through time series classification (TSC) using the deep learning framework LSTM-FCN. A TSC classifier is learned using the data (cable force in Scenario 1 and cable force ratio in Scenario 2) collected in the intact conditions. When tested on the new data with possible damage, the classifier yields the least classification accuracy on the time series collected from the most probable damaged cable or cable pair. Combining the results of the two scenarios, the most probable damaged cable is detected, which is consistent with the results in the literature. This study proposes a data-driven methodology for cable damage detection that requires the least work of data preprocessing and feature engineering, which enables fast and convenient early detection and warning in real applications. More case studies will be conducted in future work.
	
	\section*{ACKNOWLEDGMENTS}
	The authors would like to acknowledge the organizing committee of the 1st IPC-SHM 2020 for providing the data used in this paper.
	
	\bibliography{P3}

\begin{thebibliography}{36}
\providecommand{\natexlab}[1]{#1}
\providecommand{\url}[1]{\texttt{#1}}
\expandafter\ifx\csname urlstyle\endcsname\relax
  \providecommand{\doi}[1]{doi: #1}\else
  \providecommand{\doi}{doi: \begingroup \urlstyle{rm}\Url}\fi

\bibitem[Li et~al.(2018)Li, Wei, Bao, and Li]{li2018condition}
Shunlong Li, Shiyin Wei, Yuequan Bao, and Hui Li.
\newblock Condition assessment of cables by pattern recognition of
  vehicle-induced cable tension ratio.
\newblock \emph{Engineering Structures}, 155:\penalty0 1--15, 2018.

\bibitem[Li and Ou(2016)]{li2016state}
Hui Li and Jinping Ou.
\newblock The state of the art in structural health monitoring of cable-stayed
  bridges.
\newblock \emph{Journal of Civil Structural Health Monitoring}, 6\penalty0
  (1):\penalty0 43--67, 2016.

\bibitem[Brownjohn et~al.(2011)Brownjohn, De~Stefano, Xu, Wenzel, and
  Aktan]{brownjohn2011vibration}
James~MW Brownjohn, Alessandro De~Stefano, You-Lin Xu, Helmut Wenzel, and
  A~Emin Aktan.
\newblock Vibration-based monitoring of civil infrastructure: challenges and
  successes.
\newblock \emph{Journal of Civil Structural Health Monitoring}, 1\penalty0
  (3-4):\penalty0 79--95, 2011.

\bibitem[Ko and Ni(2005)]{ko2005technology}
JM~Ko and Yi~Qing Ni.
\newblock Technology developments in structural health monitoring of
  large-scale bridges.
\newblock \emph{Engineering structures}, 27\penalty0 (12):\penalty0 1715--1725,
  2005.

\bibitem[Macdonald and Daniell(2005)]{macdonald2005variation}
John~HG Macdonald and Wendy~E Daniell.
\newblock Variation of modal parameters of a cable-stayed bridge identified
  from ambient vibration measurements and fe modelling.
\newblock \emph{Engineering Structures}, 27\penalty0 (13):\penalty0 1916--1930,
  2005.

\bibitem[Miyashita and Nagai(2008)]{miyashita2008vibration}
Takeshi Miyashita and Masatsugu Nagai.
\newblock Vibration-based structural health monitoring for bridges using laser
  doppler vibrometers and mems-based technologies.
\newblock \emph{Int. J. Steel Struct}, 8\penalty0 (4):\penalty0 325--331, 2008.

\bibitem[Li et~al.(2014{\natexlab{a}})Li, Huang, Wang, and Lei]{li2014design}
ShiBin Li, Wei Huang, ZhenGuo Wang, and Jing Lei.
\newblock Design and aerodynamic investigation of a parallel vehicle on a
  wide-speed range.
\newblock \emph{Science China Information Sciences}, 57\penalty0 (12):\penalty0
  1--10, 2014{\natexlab{a}}.

\bibitem[Yang et~al.(2016)Yang, Li, Nagarajaiah, Li, and Zhou]{yang2016real}
Yongchao Yang, Shunlong Li, Satish Nagarajaiah, Hui Li, and Peng Zhou.
\newblock Real-time output-only identification of time-varying cable tension
  from accelerations via complexity pursuit.
\newblock \emph{Journal of Structural Engineering}, 142\penalty0 (1):\penalty0
  04015083, 2016.

\bibitem[Li et~al.(2014{\natexlab{b}})Li, Zhang, and Jin]{li2014real}
Hui Li, Fujian Zhang, and Yizhou Jin.
\newblock Real-time identification of time-varying tension in stay cables by
  monitoring cable transversal acceleration.
\newblock \emph{Structural Control and Health Monitoring}, 21\penalty0
  (7):\penalty0 1100--1117, 2014{\natexlab{b}}.

\bibitem[Li et~al.(2009)Li, Ou, and Zhou]{li2009applications}
Hui Li, Jinping Ou, and Zhi Zhou.
\newblock Applications of optical fibre bragg gratings sensing technology-based
  smart stay cables.
\newblock \emph{Optics and Lasers in Engineering}, 47\penalty0 (10):\penalty0
  1077--1084, 2009.

\bibitem[Bao et~al.(2019)Bao, Chen, Wei, Xu, Tang, and Li]{bao2019state}
Yuequan Bao, Zhicheng Chen, Shiyin Wei, Yang Xu, Zhiyi Tang, and Hui Li.
\newblock The state of the art of data science and engineering in structural
  health monitoring.
\newblock \emph{Engineering}, 5\penalty0 (2):\penalty0 234--242, 2019.

\bibitem[Sun et~al.(2020)Sun, Shang, Xia, Bhowmick, and
  Nagarajaiah]{sun2020review}
Limin Sun, Zhiqiang Shang, Ye~Xia, Sutanu Bhowmick, and Satish Nagarajaiah.
\newblock Review of bridge structural health monitoring aided by big data and
  artificial intelligence: From condition assessment to damage detection.
\newblock \emph{Journal of Structural Engineering}, 146\penalty0 (5):\penalty0
  04020073, 2020.

\bibitem[Abdeljaber et~al.(2018)Abdeljaber, Avci, Kiranyaz, Boashash, Sodano,
  and Inman]{abdeljaber20181}
Osama Abdeljaber, Onur Avci, Mustafa~Serkan Kiranyaz, Boualem Boashash, Henry
  Sodano, and Daniel~J Inman.
\newblock 1-d cnns for structural damage detection: Verification on a
  structural health monitoring benchmark data.
\newblock \emph{Neurocomputing}, 275:\penalty0 1308--1317, 2018.

\bibitem[Wang and Chan(2009)]{wang2009review}
Liang Wang and Tommy~HT Chan.
\newblock Review of vibration-based damage detection and condition assessment
  of bridge structures using structural health monitoring.
\newblock QUT Conference Proceedings, 2009.

\bibitem[Lin et~al.(2017)Lin, Nie, and Ma]{lin2017structural}
Yi-zhou Lin, Zhen-hua Nie, and Hong-wei Ma.
\newblock Structural damage detection with automatic feature-extraction through
  deep learning.
\newblock \emph{Computer-Aided Civil and Infrastructure Engineering},
  32\penalty0 (12):\penalty0 1025--1046, 2017.

\bibitem[Zheng et~al.(2017)Zheng, Ristovski, Farahat, and Gupta]{zheng2017long}
Shuai Zheng, Kosta Ristovski, Ahmed Farahat, and Chetan Gupta.
\newblock Long short-term memory network for remaining useful life estimation.
\newblock In \emph{2017 IEEE international conference on prognostics and health
  management (ICPHM)}, pages 88--95. IEEE, 2017.

\bibitem[Guo et~al.(2017)Guo, Li, Jia, Lei, and Lin]{guo2017recurrent}
Liang Guo, Naipeng Li, Feng Jia, Yaguo Lei, and Jing Lin.
\newblock A recurrent neural network based health indicator for remaining
  useful life prediction of bearings.
\newblock \emph{Neurocomputing}, 240:\penalty0 98--109, 2017.

\bibitem[Zhang et~al.(2020)Zhang, Liu, and Sun]{zhang2020physics}
Ruiyang Zhang, Yang Liu, and Hao Sun.
\newblock Physics-informed multi-lstm networks for metamodeling of nonlinear
  structures.
\newblock \emph{arXiv preprint arXiv:2002.10253}, 2020.

\bibitem[Abdelfattah et~al.(2018)Abdelfattah, Abdelrahman, and
  Wang]{abdelfattah2018augmenting}
Sherif~M Abdelfattah, Ghodai~M Abdelrahman, and Min Wang.
\newblock Augmenting the size of eeg datasets using generative adversarial
  networks.
\newblock In \emph{2018 International Joint Conference on Neural Networks
  (IJCNN)}, pages 1--6. IEEE, 2018.

\bibitem[Ma et~al.(2018)Ma, Xiao, and Wang]{ma2018health}
Tengfei Ma, Cao Xiao, and Fei Wang.
\newblock Health-atm: A deep architecture for multifaceted patient health
  record representation and risk prediction.
\newblock In \emph{Proceedings of the 2018 SIAM International Conference on
  Data Mining}, pages 261--269. SIAM, 2018.

\bibitem[Fawaz et~al.(2018)Fawaz, Forestier, Weber, Idoumghar, and
  Muller]{fawaz2018evaluating}
Hassan~Ismail Fawaz, Germain Forestier, Jonathan Weber, Lhassane Idoumghar, and
  Pierre-Alain Muller.
\newblock Evaluating surgical skills from kinematic data using convolutional
  neural networks.
\newblock In \emph{International Conference on Medical Image Computing and
  Computer-Assisted Intervention}, pages 214--221. Springer, 2018.

\bibitem[Higdon et~al.(2019)Higdon, El~Mokhtari, and
  Ba{\c{s}}ar]{higdon2019time}
Ben~Peachey Higdon, Karim El~Mokhtari, and Ay{\c{s}}e Ba{\c{s}}ar.
\newblock Time-series-based classification of financial forecasting
  discrepancies.
\newblock In \emph{International Conference on Innovative Techniques and
  Applications of Artificial Intelligence}, pages 474--479. Springer, 2019.

\bibitem[Chao et~al.(2019)Chao, Zhipeng, and Yuanjie]{chao2019novel}
Luo Chao, Jiang Zhipeng, and Zheng Yuanjie.
\newblock A novel reconstructed training-set svm with roulette cooperative
  coevolution for financial time series classification.
\newblock \emph{Expert Systems with Applications}, 123:\penalty0 283--298,
  2019.

\bibitem[Gao et~al.(2019)Gao, Li, Hong, and Long]{gao2019day}
Mingming Gao, Jianjing Li, Feng Hong, and Dongteng Long.
\newblock Day-ahead power forecasting in a large-scale photovoltaic plant based
  on weather classification using lstm.
\newblock \emph{Energy}, 187:\penalty0 115838, 2019.

\bibitem[Orsenigo and Vercellis(2010)]{orsenigo2010combining}
Carlotta Orsenigo and Carlo Vercellis.
\newblock Combining discrete svm and fixed cardinality warping distances for
  multivariate time series classification.
\newblock \emph{Pattern Recognition}, 43\penalty0 (11):\penalty0 3787--3794,
  2010.

\bibitem[Seto et~al.(2015)Seto, Zhang, and Zhou]{seto2015multivariate}
Skyler Seto, Wenyu Zhang, and Yichen Zhou.
\newblock Multivariate time series classification using dynamic time warping
  template selection for human activity recognition.
\newblock In \emph{2015 IEEE Symposium Series on Computational Intelligence},
  pages 1399--1406. IEEE, 2015.

\bibitem[Xing et~al.()Xing, Pei, and Keogh]{xingbrief}
Z~Xing, J~Pei, and E~Keogh.
\newblock A brief survey on sequence classification. acm sigkdd explor. newsl.
  12 (1), 40--48 (2010).

\bibitem[Fawaz et~al.(2019)Fawaz, Forestier, Weber, Idoumghar, and
  Muller]{fawaz2019deep}
Hassan~Ismail Fawaz, Germain Forestier, Jonathan Weber, Lhassane Idoumghar, and
  Pierre-Alain Muller.
\newblock Deep learning for time series classification: a review.
\newblock \emph{Data Mining and Knowledge Discovery}, 33\penalty0 (4):\penalty0
  917--963, 2019.

\bibitem[Zheng et~al.(2014)Zheng, Liu, Chen, Ge, and Zhao]{zheng2014time}
Yi~Zheng, Qi~Liu, Enhong Chen, Yong Ge, and J~Leon Zhao.
\newblock Time series classification using multi-channels deep convolutional
  neural networks.
\newblock In \emph{International Conference on Web-Age Information Management},
  pages 298--310. Springer, 2014.

\bibitem[Santos and Kern(2016)]{santos2016literature}
Tiago Santos and Roman Kern.
\newblock A literature survey of early time series classification and deep
  learning.
\newblock In \emph{Sami@ iknow}, 2016.

\bibitem[L{\"a}ngkvist et~al.(2014)L{\"a}ngkvist, Karlsson, and
  Loutfi]{langkvist2014review}
Martin L{\"a}ngkvist, Lars Karlsson, and Amy Loutfi.
\newblock A review of unsupervised feature learning and deep learning for
  time-series modeling.
\newblock \emph{Pattern Recognition Letters}, 42:\penalty0 11--24, 2014.

\bibitem[Karim et~al.(2017)Karim, Majumdar, Darabi, and Chen]{karim2017lstm}
Fazle Karim, Somshubra Majumdar, Houshang Darabi, and Shun Chen.
\newblock Lstm fully convolutional networks for time series classification.
\newblock \emph{IEEE access}, 6:\penalty0 1662--1669, 2017.

\bibitem[Karim et~al.(2019{\natexlab{a}})Karim, Majumdar, and
  Darabi]{karim2019insights}
Fazle Karim, Somshubra Majumdar, and Houshang Darabi.
\newblock Insights into lstm fully convolutional networks for time series
  classification.
\newblock \emph{IEEE Access}, 7:\penalty0 67718--67725, 2019{\natexlab{a}}.

\bibitem[Karim et~al.(2019{\natexlab{b}})Karim, Majumdar, Darabi, and
  Harford]{karim2019multivariate}
Fazle Karim, Somshubra Majumdar, Houshang Darabi, and Samuel Harford.
\newblock Multivariate lstm-fcns for time series classification.
\newblock \emph{Neural Networks}, 116:\penalty0 237--245, 2019{\natexlab{b}}.

\bibitem[Dau et~al.(2019)Dau, Bagnall, Kamgar, Yeh, Zhu, Gharghabi,
  Ratanamahatana, and Keogh]{dau2019ucr}
Hoang~Anh Dau, Anthony Bagnall, Kaveh Kamgar, Chin-Chia~Michael Yeh, Yan Zhu,
  Shaghayegh Gharghabi, Chotirat~Annh Ratanamahatana, and Eamonn Keogh.
\newblock The ucr time series archive.
\newblock \emph{IEEE/CAA Journal of Automatica Sinica}, 6\penalty0
  (6):\penalty0 1293--1305, 2019.

\bibitem[He et~al.(2015)He, Zhang, Ren, and Sun]{he2015delving}
Kaiming He, Xiangyu Zhang, Shaoqing Ren, and Jian Sun.
\newblock Delving deep into rectifiers: Surpassing human-level performance on
  imagenet classification.
\newblock In \emph{Proceedings of the IEEE international conference on computer
  vision}, pages 1026--1034, 2015.

\end{thebibliography}

\end{document}